# Artificial Intelligence Techniques for Steam Generator Modelling

Sarah Wright and Tshilidzi Marwala

*Abstract* — This paper investigates the use of different Artificial Intelligence methods to predict the values of several continuous variables from a Steam Generator. The objective was to determine how the different artificial intelligence methods performed in making predictions on the given dataset. The artificial intelligence methods evaluated were Neural Networks, Support Vector Machines, and Adaptive Neuro-Fuzzy Inference Systems. The types of neural networks investigated were Multi-Layer Perceptions, and Radial Basis Function. Bayesian and committee techniques were applied to these neural networks. Each of the AI methods considered was simulated in Matlab. The results of the simulations showed that all the AI methods were capable of predicting the Steam Generator data reasonably accurately. However, the Adaptive Neuro-Fuzzy Inference system out performed the other methods in terms of accuracy and ease of implementation, while still achieving a fast execution time as well as a reasonable training time.

## I. INTRODUCTION

Artificial Intelligence (AI) methods are concerned with machines or computer systems that have the ability to "learn" and solve problems, and as a result exhibit "intelligent" behaviour. Normally, intelligent behaviour is associated with characteristics such as having the ability to adapt, learn new skills, and form complex relationships [1]. There are several artificial intelligence methods that have been developed such as Neural Networks, Support Vector Machines, and Neuro-Fuzzy Systems. These AI systems have been utilised in different applications for example: pattern recognition, prediction of process variables, and various control applications. Each of these methods has different approaches to adapting and learning in order to emulate intelligent behaviour. Such Artificial Intelligence methods are particularly useful in modelling complex relationships where the relationship cannot be computed directly or easily interpreted by a human.

A well known Artificial Intelligence method is Neural

School of Electrical and Information Engineering
P/Bag x3, Wits, 2050 South Africa
Website: www.tshilidzimarwala.com

Networks. Neural networks are inspired by the mechanisms of the human brain and are capable of learning complex relationships through the association of examples of these relationships [2]. A neural network continuously adapts or adjusts these complex relationships found in an example dataset until it has learnt the relationships sufficiently well. Neural networks model these complex relationships in terms of a set of free parameters (weights and biases) that can be mathematically represented by a function [3]. There are numerous types of neural networks that can be implemented such as Radial Basis Function and Multi-layer Perceptions.

Support Vector Machines is a more recent Artificial Intelligence method developed by Vapnik and his colleges' in 1992. Support Vector Machines are based on statistical learning where the goal is to determine an unknown dependency between a set of inputs and outputs, and this dependency is estimated from a limited set of example data [4]. In the case of classification, the idea is to construct a hyper-plane as a decision surface in such a way that the margin of separation between the different classes is maximized [5]. In the class of function approximation, the goal is to find a function that has at most a certain deviation from the desired target for all the points in a dataset of examples used to model such dependencies. Like neural networks, it models complex relationships using a mathematical approach.

Neuro-Fuzzy Systems are based on Fuzzy logic which was formulated in the 1960s by Zadeh [6]. These systems combine Fuzzy Logic and certain principles of Neural Networks in order to model complex relationships. Fuzzy systems use a more linguistic approach rather than a mathematical approach, where relationships are described in natural language using linguistic variables [6].

All of the AI methods mentioned require a dataset in order to train the AI systems to be able to model the complex relationships of the system being modelled. Therefore, the AI system learns by example through a training process, and this dataset is called a training dataset.

Most AI methods can be used for function approximation in which predictions of continuous variables can be generated. In this paper, the investigation into certain AI methods for the application of predicting certain variables from a Steam Generator will be discussed. The AI methods that were investigated were: Neural Networks (Radial Basis

Function, Multi-Layer Perception, Committees, and Bayesian Techniques), Support Vector Machines, and Adaptive Neuro-Fuzzy Inference Systems. Each of theses AI methods were investigated and simulated in Matlab, in order to ascertain the performance of each method as well as its strengths and weakness when applied to the stated application. The main performance measures under consideration are the accuracy obtained, speed of training, and the speed of execution of the AI system on unseen data.

The paper will first give a basic foundation of the theory of the AI methods used, and then the implementations and their results will be presented. Finally, the key findings of the simulations will be discussed.

## II. THE STEAM GENERATOR DATASET

The problem requires modelling the input-output relationship of data obtained from a Steam Generator at Abbott Power Plant in the USA. The dataset used is available online, and contains 9600 samples. The model consists of 4 inputs and 4 outputs. The inputs are the input fuel, air, reference level (inches), and disturbance defined by the load level. The fuel and air inputs have been scaled between 0 and 1. The outputs are the drum pressure (PSI), excess oxygen in exhaust gases (percentage), the level of water in the drum, and the steam flow (Kg/s). Both the inputs and outputs are in numeric form and have different units to express their quantities. The idea is to be able to predict the outputs for a specific set of inputs for the steam generator. Therefore, the problem is a regression problem as the goal is to predict the value of a number of continuous variables. This data set will be modelled using the different artificial intelligence methods mentioned above.

## III. ARTIFICIAL NEURAL NETWORKS

### A. General Theory of Artificial Neural Networks

Neural Networks were originally inspired by the mechanisms used by the human brain to learn by experience and processes information. The human brain consists of many interconnected neurons that form an information processing network capable of learning and adapting from experience [2, 7].

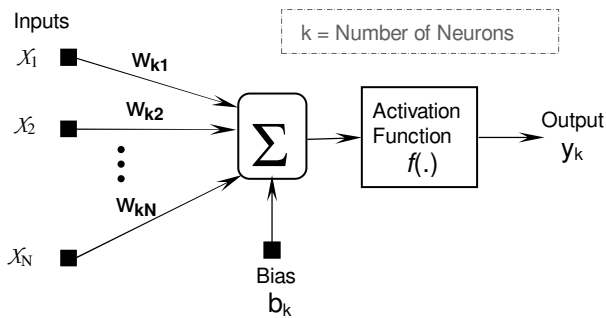

*Figure 1: Structure of a Generalised Artificial Neuron [5]*

Basically, a neural network is a data modelling technique used to form an input/output relationship for a specific problem [2]. Therefore, an input/output relationship must exist for the neural network to function in predicting outputs or classifying data. The basic component of a neural network is an artificial neuron which is a largely simplified representation of a biological neuron. Each neuron receives a number of inputs which may be from the outputs of other neurons or the source data being fed into the network [2, 7]. Each input is multiplied by a weight to determine its influence or strength. These weights are analogous to the adjustment of the synaptic connections between neurons that occurs during the learning process in biological systems [2, 7]. The weighted inputs and an external bias value are summed. The summed signal is passed through an activation function to produce the neuron's output signal. The bias value has the effect of increasing or decreasing the signal input passed to the activation function and is similar to the firing threshold of a biological neuron [2, 7]. The activation function limits the amplitude range of the neuron's output signal [5].

An artificial neuron represents the basic information processing unit of any neural network. However, the general characteristics of the artificial neuron: activation function used, biases, method of calculating the weights, and number of inputs; will differ depending on the type of neural network and the problem being modelled. Figure 1 shows the model of an artificial neuron [5]. The general mathematical model of a neuron can be described by Equation 1 [5].

$$y_k = f\left(\sum_{j=1}^{N} w_{kj} x_j + b_k\right) \quad (1)$$

where:

$y_k$ = output of the kth neuron
$x_j$ = the jth input
$w_{kj}$ = weighting connecting the jth input to neuron k.

A neural network consists of several layers of interconnected artificial neurons working together to model a problem. The types of neural networks that will be discussed are feed-forward neural networks, meaning that the signals can only travel in one direction through the network structure: from the inputs towards the outputs. The input layer consists of several inputs (source data) to be modelled by the network, it does not have any neurons and no computation is performed [5]. The network may have several hidden layers that introduce more adjustable weightings to the network, allowing a higher order model of the data to be extracted [5, 3]. The final layer is the output layer of the network which produces the overall network's outputs. Each layer of the network receives inputs from the previous layer of the network, and passes its outputs to the next layer. Normally, every node in a layer is connected to every other node in the following layer (meshed) [5]. The basic structure of a feed-forward network can be seen in Figure 2 [5].

A neural network learns by example through training algorithms. Training results in an input/output relationship being determined for a specific problem. Training can be supervised or unsupervised. The neural networks discussed will use supervised training. Supervised training involves having a training dataset where numerous examples of inputs and their corresponding outputs (targets) are fed to the network. The weights and biases of the neural network are continuously adjusted to minimise the error between the network's outputs and the target outputs [2, 5, 7].

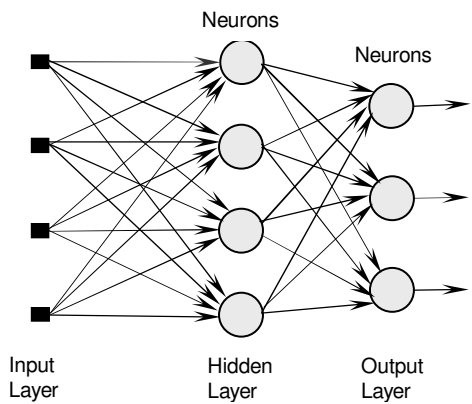

Figure 2: Basic Structure of a Feed-forward Artificial Neural Network [5]

Optimisation techniques can be used to determine the weights of the network, since it is a minimisation problem. Therefore, the knowledge or information about the problem is contained by the weights and biases of the network.

An important property of neural networks is their ability to generalise. Generalisation refers to the ability of the neural network to predict or produce reasonable outputs for inputs not seen during the training or learning process [5]. Thus, the input/output relationship computed is valid for unseen data. Generalisation is influenced by the size of the training dataset and the architecture of the neural network [5]. The best generalisation is normally achieved when the number of free parameters is fairly small compared to the size of the dataset [3]. However, a neural network can have poor generalisation if it is under-trained (under-fitting) or over-trained (over-fitting). Over-training occurs when the neural network fits the training data perfectly and results with a function approximation or boundary line that is not smooth but erratic in nature [3]. The network effectively memorises the data and therefore, has poor generalisation on data not in the training set. Also, a neural network can be under-trained: there are not enough free parameters to sufficiently form an input/output relationship that captures the features of the problem [3].

### B. Multi-Layer Perception

Multi Layer Perception (MLP) neural networks are a popular class of feed-forward networks (Figure 2). They were developed from the mathematical model of the neuron (Figure 1), and consist of a network of neurons or perceptions [2]. An MLP network consists of an input layer (source data), several hidden layers of neurons, and an output layer of neurons. The hidden layers and the output layer can have different activation functions. There are various types of activation functions that can be employed. The activation function of the hidden neurons must be nonlinear and are usually functions that are differentiable [3]. Typically, the hyperbolic tangent or logistic functions are used for the activation function of the hidden neurons. However, the output activation function may be linear. Certain activation functions are more appropriate for different types of problems, therefore, the activation function needs to be selected according to the problem. Normally, a linear output activation function is used for regression problems as it does not limit the range of the output signal [2].

A multi-layer perception neural network represents a multivariate non-linear function mapping between a set of input and output variables [3]. It has been shown that any continuous function can be modelled accurately with one hidden layer, provided there is a sufficient number of hidden neurons [3, 5]. An MLP network with one hidden layer can be mathematically represented by Equation 2 [3].

$$y_k = f\left( \sum_{j=1}^{Nhidden} w_{kj}^{(2)} f_A\left( \sum_{i=1}^{Ninput} w_{ji}^{(1)} x_i + w_{j0}^{(1)} \right) + w_{k0}^{(2)} \right) \quad (2)$$

where:

- k = number of outputs
- $y_k$ = the output at the $k^{th}$ node
- j = number of hidden neurons
- i = number of inputs
- $f_A$ = activation function of the hidden neurons
- f = activation function of the output neurons
- $x_i$ = the input from the $i^{th}$ input node
- $w_{ji}$ = weights connecting the input with the hidden nodes
- $w_{jk}$ = weights connecting the hidden with the output nodes

$w_{0j}$ and $w_{0k}$ = biases

The complexity of the model is related to the number of hidden units, as the number of free parameters (weights and biases) available to adjust is directly proportional to the number of hidden units.

Training involves continuously adjusting the values of the weights and biases to minimise the error between the network's output and the desired targets. Initially, the weights and biases are set to random values, and then adjusted using an optimisation technique. However, such optimisation techniques are highly susceptible to finding local minima, and there is no guarantee that a global minimum has been found [3]. The best way to try and avoid a solution that is a local minimum is to train many networks taking the best network produced.

## C. Radial Basis Functions

Radial Basis Functions are two-layer feed-forward neural networks with the activation function of the hidden units being radial basis functions [5]. The response of the hidden layer unit is dependent on the distance an input is from the centre represented by the radial basis function (Euclidean Distance) [2]. Each radial function has two parameters: a centre and a width. Therefore, the maximum activation of a hidden unit is achieved when the input coincides with the centre vector. The width of the basis function determines the spread of the function and how quickly the activation of the hidden node decreases with the input being an increased distance from the centre [3]. The most common radial basis function used is the Gaussian bell-shaped distribution.

Normally, an RBF only has one hidden layer, and a linear output layer. The input layer simply passes the input data to the hidden layer. An RBF network can be modelled mathematically by Equation 3 and the Gaussian activation function is represented by Equation 4. The bias parameters at the output layer compensate for the difference between mean output values and mean target values [3].

$$y_k(\mathbf{x}) = \sum_{j=1}^{M} w_{kj} \phi_j(\mathbf{x}) + w_{k0} \tag{3}$$

where:
- $y_k$ = the output at the k$^{th}$ node
- $M$ = number of hidden nodes
- $w_{kj}$ = the weight factor from the j$^{th}$ hidden node to the k$^{th}$ output node
- $w_{k0}$ = the bias parameter of the kth output node
- $\phi(x)$ = radial basis activation function

$$\phi_j(\mathbf{x}) = \exp\left\{\frac{-\|\mathbf{x} - \mathbf{u}_j\|^2}{2\sigma_j^2}\right\} \tag{4}$$

where:
- $\mathbf{x}$ = input vector
- $\mathbf{u}_j$ = centre vector of the jth hidden node
- $\sigma$ = width of basis function

An RBF is trained in two stages. The first stage is an unsupervised learning process to determine the parameters of the radial basis function for each hidden node [3]. Therefore, only the input data is used during this process. These parameters are the centres and the widths of the basis functions. There are a number of unsupervised training algorithms to determine the parameters of the basis functions such as K-means clustering. The second stage involves finding the final layer weights that minimise the error between the network's output and the target values. Therefore, the second stage is done using supervised learning. Since the output layer is a linear function, the final layer weights can be solved using linear algebra [3]. Both of these stages are relatively fast, therefore, an RBF trains much faster than an equivalent MLP. The parameters of an RBF can be determined by supervised training. However, the optimisation process is no longer linear, resulting in the process being computationally expensive compared to the two stage training process.

The main difference between MLPs and RBFs are that an MLP splits the input space into hyper-planes while an RBF splits the input space into hyper-spheres [2].

## D. Committees

Combining the outputs of several neural networks into a single solution to gain improved accuracy over an individual network output is called a committee or ensemble [8]. The simplest way of combing the outputs of different networks together is to average the outputs obtained [3]. The averaging ensemble can be expressed by Equation 5 [3, 8],

$$y_K = \frac{1}{N}\sum_{i=1}^{N} y_{Ki} \tag{5}$$

where $y_k$ is the kth output, $y_{ki}$ is the kth output of network i, and N is the number of networks in the committee. It can be shown that averaging the prediction of N networks reduces the sum-of-squares error by a factor of N [3]. However, this does not take into account that some networks in the committee may generate better predictions than others[3]. In this case, a weighted sum can be formulated in which certain networks contribute more to the final output of the committee [3]. There are several other committee methods to improve the accuracy of the prediction obtained, such as Bagging and Boosting.

## E. Bayesian Techniques for Neural Networks

The training of the neural networks using the more standard approaches relies on the minimisation of a function error (Maximum Likelihood Approach) [3]. This approach makes defining the neural network model difficult, and both training and validation datasets are necessary to determine the model that exhibits the best generalisation. There will always be a certain error between the predicted and the actual. If several networks with identical architectures are produced with the same error, the weights and biases will not be the same each time, as there is a level of uncertainty in the training process due to there being many possibilities for parameters.

In the Bayesian approach, a probability distribution function is considered to be represented over the weight space, to account for the uncertainty in determining the weight vector [3]. Instead of attempting to find a single set of weights that minimised the error between the predicted and actual values. The probability distribution represents the degree of confidence associated to the different values for the weight vector [3]. This probability distribution is initialised to some prior distribution, and then with the aid of the training dataset the posterior probability distribution can be determined and used to evaluate the predicted outputs for new input data points [3]. The posterior probability distribution can be expressed using Bayes' Theorem and is shown in Equation 6.

$$P(w|D) = \frac{P(D|w)P(w)}{P(D)} \quad (6)$$

where D represents the target values of the training dataset, w is the vector representing the adaptive weights and biases, P(w) is the probability distribution function of the weight space in absence of any data points (Prior Probability Distribution), P(D) is a normalisation factor, P(D|w) is a likelihood function, and P(w|D) is the posterior probability distribution. Using Bayes' Theorem allows any prior knowledge about the uncertain weight values to be updated based on the knowledge gained from the training dataset to produce the posterior distribution of the unknown weight values [3]. The posterior probability distribution gives an indication of which weight values for the weight vector are most probable [3].

The prior probability distribution should take into account any information known about the weights [3]. From regularisation techniques, it's known that small weight values are favoured in order to produce smooth network mappings. Therefore, the weight-decay regularisation needs to be incorporated in the prior probability distribution function. For prior probability distribution that is a Gaussian function, the form is shown in Equation 6 [3], where W is the number of weights and $Z_w$ is the normalisation coefficient. If the weight decay term is small then the p(w) is large. The quantity α is the coefficient of weight-decay.

$$P(w) = \frac{1}{Z_W(\alpha)} \exp\left(-\frac{\alpha}{2}\|w\|^2\right) \quad (6)$$

where $Z_W = \left(\frac{2\pi}{\alpha}\right)^{W/2}$

The Likelihood probability distribution is given by Equation 7, and is an expression of the difference between the predicted output ( $y(x,w)$ ) and the target output (*t*). The quantity β is the coefficient of the data error [3].

$$P(D|w) = \frac{1}{Z_D(\beta)} \exp\left(-\frac{\beta}{2}\sum_{n=1}^{N}\{y(x^n,w)-t^n\}^2\right) \quad (7)$$

where $Z_D = \left(\frac{2\pi}{\beta}\right)^{N/2}$

The posterior probability distribution can be obtained by applying Bayes' theorem and is given below [3]. It can be seen that S(w) is dependent on the sum-of-squares error function and a weight regularisation term [3].

$$P(w|D) = \frac{1}{Z_S} \exp(-S(w)) \quad (8)$$

where  $S(W) = \beta E_D + \alpha E_W$

$$= \frac{\beta}{2}\sum_{n=1}^{N}\{y(x^n,w)-t^n\}^2 + \frac{\alpha}{2}\sum_{i=1}^{W}w_i^2$$

$$Z_S(\alpha,\beta) = \int \exp(-\beta E_D - \alpha E_W)\,dw$$

The training process for the Bayesian approach involves determining the appropriate posterior probability distribution of the weight values [9]. In order, to make a prediction for a new input vector, the output distribution must be computed, and is given by Equation 9. This Equation is effectively taking an average prediction of all the models weighted by their degree of probability [3], and is dependent on the posterior probability distribution. Therefore, the trained network can make predictions on input data it has not seen by using the posterior probability distribution.

$$P(y^{n+1}|x^{n+1},D) = \int P(y^{n+1}|x^{n+1},w)P(w|D)\,dw \quad (9)$$

The evaluation of the probability distributions requires integration over a multidimensional weight space, and is not easily handled analytically. One method to evaluate the integrals is to use a Gaussian Approximation which allows the integral to be analytically evaluated using optimisation techniques [3]. Another common method used to solve these type of integrals is a random sampling method called Monte Carlo Technique [10]. Therefore, the Monte Carlo or the Hybrid Monte Carlo method is normally used to identify the posterior probability distribution of the weights for a Bayesian neural network, by sampling from the posterior weight distribution.

*F. Monte Carlo Methods*

In the Bayesian approach to neural networks, integration plays a significant role as calculations involve evaluating an integral over the weight space. Monte Carlo is a method of approximating the integral by using a sample of points from the function of interest [3]. The integrals that need to be evaluated are of the form [3],

$$I = \int F(w)P(w|D)dw \quad (10)$$

where F(w) is the integrand and P(w|D) is the posterior distribution of weights. This integral can then be approximated using a finite sum of the form,

$$I \approx \frac{1}{L}\sum_{i=1}^{L} F(w_i) \quad (11)$$

where $w_i$ is the sample of weight vectors generated from the posterior probability distribution [3].

In order to generate samples of the weight vector space representative of the P(w|D), a random search through the weight space for areas were the distribution is reasonably large is performed. This done using a technique called Markov Chain Monte Carlo, where a sequence of weight vectors are generated, each new vector in the sequence depending on the previous weight vector plus a random component [3]. A random walk is the simplest method in which each successive step is computed using Equation 12 [3].

$$w_{n+1} = w_n + \varepsilon \quad (12)$$

ε is a random vector that allows more of the weight space to be explored. In order, to find samples of weight vectors that are representative of the P(w|D) distribution, a procedure known as the Metropolis Algorithm is used to select the sample weight vectors. The Metropolis Algorithm rejects or accepts a certain sample of the weight space or state generated using Equation 12 based on the following conditions,

if $P(w_{n+1}|D) > P(w_n|D)$ accept state $w_{n+1}$
if $P(w_{n+1}|D) < P(w_n|D)$ accept state $w_{n+1}$
  with probability $\dfrac{P(w_{n+1}|D)}{P(w_n|D)}$

Using the above conditions, certain of the weight vector samples will be rejected if they lead to a reduction in the posterior distribution [3]. This procedure is repeated a number of times until the necessary number of samples are produced for the evaluation of the finite sum for the integral. Due to high correlation in the posterior distribution as a result of the each successive step being dependent on the previous, a large number of the new weight vector states will be rejected [3]. Therefore, a Hybrid Monte Carlo method can be used instead.

The Hybrid Monte Carlo methods uses information about the gradient of P(w|D) to ensure that samples through the areas of higher posterior probabilities are favoured [3]. This gradient information can be obtained through the back-propagation algorithm. The Hybrid Monte Carlo method is based on the principles of Hamiltonian mechanics that describe molecular dynamics [10]. It is a form of the Markov Chain, however, the transition between states is achieved using the stochastic dynamic model [9]. In statistical mechanics, the state space of a system at a certain time can be described by the position and momentum of all the molecules of the system at that time [9]. The position defines the potential energy of the system and the momentum defines the kinetic energy of the system [9-10]. The total energy of the system is the sum of the potential and kinetic energy, and can be represented by the Hamiltonian equation defined as,

$$H(w,p) = E(w) + K(p) = U(w) + \frac{1}{2}\sum_i p_i^2 \quad (13)$$

where w is the position variable, p is the momentum variable, H(w,p) is the total energy of the system, E(w) is the potential energy, and K(p) is the kinetic energy. The positions are analogous with the weights of a neural network, and potential energy with the network error [10]. In this equation, the energies of the system are defined by energy functions representing the state of the physical system (canonical distributions) [10]. In order to obtain the posterior distribution of the network weights, the following distribution is sampled ignoring the distribution of the momentum vector [9].

$$P(w,p) = \frac{1}{Z}\exp(-H(w,p)) \quad (14)$$

Hamiltonian dynamics are used to sample at a fixed energy in terms of a fictitious time τ [9-10], and are shown in Equation 15 and 16. Since the dynamics shown in Equations 15 and 16 can not be simulated exactly, the equations are discretised using finite time steps given by Equation 17 and 19. [10]. In this way the position and momentum at time τ + ε is expressed in terms of the position and momentum at time τ [10]. This method is known as the leap-frog method. These new states are accepted or rejected using the Metropolis criterion.

$$\frac{dw_i}{d\tau} = \frac{\partial H}{\partial p_i} = p_i \quad (15)$$

$$\frac{dp_i}{d\tau} = -\frac{\partial H}{\partial w_i} = -\frac{\partial E}{\partial w_i} \quad (16)$$

*Discretised Equations*

$$\hat{p}_i\left(\tau + \frac{\varepsilon}{2}\right) = \hat{p}_i(\tau) - \frac{\varepsilon}{2}\frac{\partial E}{\partial w_i}[\hat{w}_i(\tau)]$$
(17)

$$\hat{w}_i(\tau + \varepsilon) = \hat{w}_i(\tau) + \varepsilon\hat{p}_i\left(\tau + \frac{\varepsilon}{2}\right)$$
(18)

$$\hat{p}_i(\tau + \varepsilon) = \hat{p}_i\left(\tau + \frac{\varepsilon}{2}\right) - \frac{\varepsilon}{2}\frac{\partial E}{\partial w_i}[\hat{w}_i(\tau + \varepsilon)]$$
(19)

The basic steps in the implementation of the Hybrid Monte Carlo algorithm are [9, 11]:

(i) Randomly choose a trajectory direction (λ) where λ is -1 for a backward trajectory and +1 for a forward trajectory.

(ii) Starting from a current state (w, p). Perform L leapfrog steps with the step size ε using Equations 16-19 to product a candidate state (w*, p*). Performing L leapfrog steps allows more of the state space to be explored faster.

(iii) Using the Metropolis criterion, accept or reject the (w*, p*) state. If the candidate state is rejected the old state (w, p) is kept as the new state. Otherwise, the candidate state is accepted and it becomes the new state.

## IV. SUPPORT VECTOR MACHINES

Support Vector Machines (SVM) were introduced by Vapnik and his colleges in 1992. They are based on statistical learning theory and are one type of kernel learning algorithm in the field of machine learning [4]. SVMs can be used for both classification and regression problems. The goal of statistical learning is to determine an unknown dependency between a set of inputs and outputs, and this dependency is estimated from a limited set of example data. [4]. Therefore, the objective of a SVM, like neural networks, is to produce a model which can predict the output values of a dataset previously unseen. Thus, SVMs utilise supervised learning techniques, and require a training and testing dataset.

In the case of classification, the idea is to construct a hyper-plane as a decision surface in such a way that the margin of separation between the different classes is maximized [5]. These decision planes are defined to act as decision boundaries separating different classes of objects.

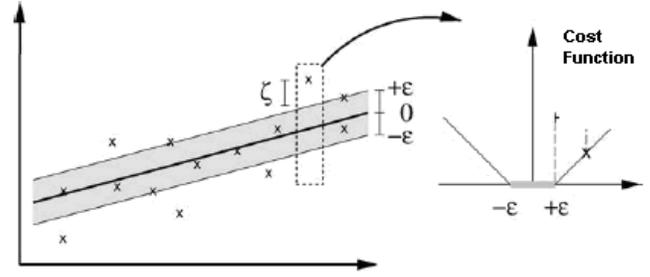

Figure 3: Showing a Linear SVM Regression for a Dataset Illustrating the $\varepsilon$-Tube and Penalty Cost Function [12]

In support vector regression, the idea is to find a function that has at most a deviation of $\varepsilon$ from the desired targets for all the training data ($\varepsilon$-SV regression) [12]. Thus, errors below the deviation are not of concern, and points outside this deviation are penalized (Refer to Figure 3). Therefore, a function that approximates all the input-output pairs with the defined precision must actually exist and the optimisation required must be able to be feasibly solved [12]. In order, to account for data points that cannot be easily modelled, slack variables are normally introduced.

In both classification and regression, the inputs are mapped into a higher dimensional feature space by a function $\phi(x)$ induced by a kernel function [4, 12]. The SVM then finds a linear separating hyper-plane with the maximal margin in this higher dimensional space for the classification case, and a set of linear functions in this higher dimensional space for the regression case [4, 13]. There are different types of kernel functions: linear, polynomial, radial basis function (RBF), and sigmoid. Any function that satisfies Mercer's Theorem can be used as a kernel function [4]. The kernel function is equal to the inner product of the two vectors (input vector ($\mathbf{x}$) and input pattern of the ith training sample ($\mathbf{x_i}$)) induced in the feature space and is given by Equation 20 [5].

$$K(x, x_i) = \phi^T(x)\phi(x_i)$$
(20)

In the case of regression, if given a training dataset, $\{(x_i, t_i)\}_{i=1}^{N}$, where the $x$ is the input vector and $t$ is the target value, an SVM approximates the function using Equation 21 [13].

$$y = f(x) = w\phi(x) + b$$
(21)

where $\phi(x)$ represents the higher dimensional feature space that the inputs are mapped to, w is the weight vector, and b is the bias. Since in reality, not every point will be able to fit within the deviation defined, the Support Vector Machine minimises the number of points outside the deviation using a penalty parameter [12]. This is achieved by minimising Equation 22. If Equation 22 is transformed into dual formulation, it is expressed in terms of the kernel function and support vectors. Support vectors consist of the data points that sit on the boundaries of the acceptable region defined by $\varepsilon$ and are extracted from the training dataset [5, 13]. This constrained optimisation problem can be solved using quadratic programming with the training data and, as a result, is guaranteed to find a global optimum [5, 12].

*Minimise:*

$$\frac{1}{2}\|w\|^2 + C\sum_{i=1}^{N}(\xi_i + \xi_i^*)$$
(22)

*Subject to:*

$$t_i - w\phi(x_i) - b \leq \varepsilon + \xi_i$$
$$w\phi(x_i) + b - t_i \leq \varepsilon + \xi_i^*$$
$$\xi_i, \xi_i^* \geq 0$$

where  w -  weight vector
       C -  penalty parameter
       N -  number of data points
       $\varepsilon$ -  deviation from function
       $\xi_i, \xi_i^*$ - Slack Variables

The constraints above deal with a linear $\varepsilon$-insensitive loss function used to penalise the data points in the training dataset that are outside the specified deviation $\varepsilon$. A loss function is used to determine which function (f(x)) best describes the dependency observed in the training dataset [4]. The purpose of the loss function is to determine the cost of the difference between the actual and predicted outputs for a given set of inputs. The $\varepsilon$-insensitive loss function is defined by Equation 23 [4, 5]. As seen in Figure 3, in regression problems a $\varepsilon$-tube is formed around the function, and any data points outside this $\varepsilon$-tube have an associated cost given by Equation 23. Most data points should lay within the $\varepsilon$-tube, however, the slack variable allow some data points to lie outside the $\varepsilon$-tube [5]. There are two slack variables to account for the upper and lower bounds of the $\varepsilon$-tube. Both $\varepsilon$ and C are user-defined parameters. The parameter C is a regularisation parameter that controls the trade-off between the complexity of the machine and the number of data points that lie outside the $\varepsilon$-tube [5]. The deviation $\varepsilon$, determines the approximation accuracy enforced on the training data points [13]. For regression, the parameters C and $\varepsilon$ should be tuned simultaneously [5].

$$\ell(f(x),t) = \begin{cases} |f(x)-t|-\varepsilon & \text{if } |f(x)-t| > \varepsilon \\ 0 & \text{otherwise} \end{cases}$$
(23)

There are other $\varepsilon$-insensitive loss functions that can be used such as a quadratic $\varepsilon$-insensitive loss function. Also, a least squares cost function can be used. This results in a Least Squares Support Vector Machine (LS-SVM) that has a few different properties to the original Vapnik's SVM presented above.

In a least squares SVM, the $\varepsilon$-insensitive loss function is replaced by a least squares cost function which corresponds to a form of ridge regression [14]. The inequality constraints that Equation 22 is subject to are replaced by equality. As a consequence, the training process of a LS-SVM involves solving a set of linear equations instead of a quadratic programming problem. The set of linear equations that result are of the dimension N+1, where N is the number of training samples [15]. In the case of a standard SVM, the quadratic programming (QP) problem to be solved is roughly exponential to the size of the training dataset [14]. Therefore, the number of training samples used to train an SVM should be considered carefully. However, the set of linear equations is still not as time and computationally consuming to solve as the QP problem.

In a LS-SVM, the weight vector that results from minimizing the summed squared approximation error over all training samples is searched for, where the approximation error is the difference between the SVM's output and the desired target output [15]. The equation for a LS-SVM is shown below in Equation 24 [15].

*Minimise:*

$$\frac{1}{2}\|w\|^2 + C\sum_{i=0}^{N} e_i^2 \qquad (24)$$

*Subject to:*

$$t_i = w\phi(x_i) + b + e_i$$

where $e_i = t_i - f(x_i)$

The main difference between Neural networks and Support Vector Machines is that support vector machines minimise an upper bound of the generalisation errors instead of minimising the error on the training dataset [13]. Support vector machines utilise risk minimisation, measured using a loss function. Normally, support vector machines have a slower execution time as there is little control over the number of support vectors defined [5]. An SVM has less parameters to tune than a neural network, and the optimisation procedure can be performed efficiently. In the case of the LS-SVM, the parameters that need to be tuned are the penalty or regularisation constant and the deviation of the Gaussian function, if an RBF kernel is used. While for a standard SVM the $\varepsilon$ accuracy for the $\varepsilon$-insensitive loss function needs to determined additionally.

## V. FUZZY LOGIC AND NEURO-FUZZY SYSTEMS

Neuro-Fuzzy Systems are based on Fuzzy logic which was formulated in the 1960s by Zadeh. These systems combine Fuzzy Logic and certain principles of Neural Networks in order to model complex relationships. Fuzzy systems use a more linguistic approach rather than a mathematical approach, where relationships are described in natural language using linguistic variables. Fuzzy Logic can deal with ill-defined, imprecise systems [16], and therefore are a good tool for system modelling. This section introduces the basics of Fuzzy Logic and then explains Adaptive Neuro-Fuzzy Inference Systems that are based on the foundations of Fuzzy Logic.

### A. Basic Fuzzy Logic Theory

Fuzzy logic is a method of mapping an input space to an output space by means of a list of linguistic rules that consist of if-then statements [6]. Fuzzy logic consists of 4 components: fuzzy sets, membership functions, fuzzy logical operators, and fuzzy rules [6, 17, 18].

In classical set theory, an object is either a member or is not a member of specific set [17-18]. Therefore, it is possible to determine if an object belongs to a specific set as a set has clear distinct boundaries, provided an object cannot achieve partial membership. Another way of thinking about this is that the object's belonging to a set is either true or false. A characteristic function for a classical set has a value of 1 if the object belongs to the set and a value of zero if the object doesn't belong to the set [17]. For example, if a set X is defined to represent all possible heights of people, one could define a "tall" subset for any person who is above or equal to a specific height $x$, and anyone below $x$ doesn't belong to the "tall" set but to a "short" subset. This is clearly inflexible as a person just below the boundary is labelled as being short when they are clearly tall to some degree. Therefore, intermediate values such as fairly tall are not allowed. Also, these clear cut defined boundaries can be very subjective in terms of what a person may define as belonging to a specific set.

The main aim behind fuzzy logic is to allow a more flexible representation of sets of objects by using a fuzzy set. A fuzzy set does not have as clear cut boundaries as a classical set, and the objects are characterized by a degree of membership to a specific set [17-18]. Therefore, intermediate values of objects can be represented which is closer to the way the human brain thinks opposed to the clear cut-off boundaries in classical sets. A membership function defines the degree that an object belongs to a certain set or class. The membership function is a curve that maps the input space variable to a number between 0 and 1, representing the degree that a specific input variable belongs to a specific set [17-18]. A membership function can be a curve of any shape. Using the example above, there would be two subsets one for tall and one for short that would overlap. In this way a person can have a partial participation in each of these sets, therefore, determining the degree to which the person is both tall and short.

Logical operators are defined to generate new fuzzy sets from the existing fuzzy sets. In classical set theory there are 3 main operators used, allowing logical expressions to be defined: intersection, union, and the complement [17]. These operators are used in fuzzy logic, and have been adapted to deal with partial memberships. The intersection (AND operator) of two fuzzy sets is given by a minimum operation, and the union (OR operator) of two fuzzy sets is given by a maximum operation [17]. These logical operators are used in the rules and determination of the final output fuzzy set.

Fuzzy Rules formulate the conditional statements which are used to model the input-output relationships of the system, and are expressed in natural language [6]. These linguistic rules are in the form of if-then statements which use the logical operators and membership functions to produce an output. An important property of fuzzy logic is the use of linguistic variables. Linguistic variables are variables that take words or sentences as their values instead of numbers [17]. Each linguistic variable takes a linguistic value that corresponds to a fuzzy set [17], and the set of values that it can take is called the term set [18]. For example, a linguistic variable *Height* could have the following term set *{very tall, tall, medium, short, very short}*. A single fuzzy rule is of the form:

*if x is A then y is B* (25)

where A and B are fuzzy sets defined for the input and output space respectively. Both x and y are linguistic variables, while A and B are the linguistic values or labels represented by the membership functions [16]. Each rule consists of two parts: the antecedent and the consequent [17]. The antecedent is the component of the rule falling between the if-then, and maps the input x to the fuzzy set A, using a membership function. The consequent is the component of the rule after the then, and maps the output y to a membership function. The input membership values act like weighting factors to determine their influence on the fuzzy output sets [17]. A fuzzy system consists of a list of these if-then rules which are evaluated in parallel. The antecedent can have more than one linguistic variable, these inputs are combined using the AND operator.

Each of the rules is evaluated for an input set, and corresponding output for the rule obtained. If an input corresponds to two linguistic variable values then the rules associated with both these values will be evaluated. Also, the rest of the rules will evaluated, however, will not have an effect on the final result as the linguistic variable will have a value of zero. Therefore, if the antecedent is true to some degree, the consequent will have to be true to some degree [17]. The degree of each linguistic output value is then computed by performing a combined logical sum for each membership function [17]. After which all the combined sums for a specific linguistic variable can be aggregated. These last stages involve the use of an inference method which will map the result onto an output membership function [19]. Finally, a defuzzification process is preformed in which a single numeric output produced. One method of computing the degree of each linguistic output value is to take the maximum of all rules describing this linguistic output value [17, 19], and the output is taken as the centre of gravity of the area under the effected part of the output membership function. There are other inference methods such as averaging and sum mean square [19]. Figure 4 shows the steps involved in creating an input-output mapping using fuzzy logic [20].

The use of a series of fuzzy rules, and inference methods to produce a defuzzified output constitute a Fuzzy Inference System (FIS) [21]. The final manner in which the aggregation process takes place and the method of defuzzification can differ depending on the implementation of the FIS chosen. The approach discussed above is that of the Mamdani based FIS.

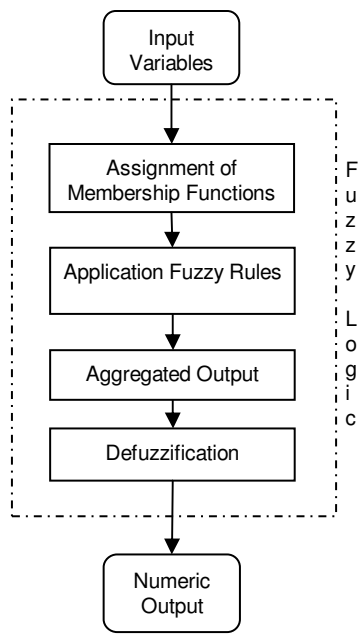

*Figure 4: Showing the Steps Involved in the Application of Fuzzy Logic to a Problem [20]*

There are several types of fuzzy inference systems which vary according to the fuzzy reasoning and the form of the if-then statements applied [16]. Another method of Fuzzy inference that is worth discussing is the Takagi-Sugeno-Kang method. It is similar to the Mamdani approach described above except that the consequent part is of a different form and as a result the defuzzification procedure is different. The if-then statement of a Sugeno fuzzy system expresses the output of each rule as a function of the input variables, and has the form [1],

*if* x is A  *AND*  y is B *then* z =  f(x,y)  (26)

If the output of each rule is a linear combination of the input variables plus a constant, then it is known as a first-order Segeno fuzzy model, and has the form [1]:

$$z = px + qy + c \qquad (27)$$

Alternatively, the output of a rule can be a constant. The final output of the Sugeno FIS is a weighted average of the outputs from each rule [16].

### B. Adaptive Neuro-Fuzzy Inference Systems

The main difficulties with Fuzzy Inference Systems are that it is difficult to transform human knowledge into the necessary rule base, as well as to adjust the Membership functions to achieve a minimized output error for the FIS [16]. The purpose of an ANFIS (Adaptive Neuro-Fuzzy Inference System *or* Adaptive Network-Based Fuzzy Inference System) is to establish a set of rules along with a set of suitable membership functions that is capable of representing the input/output relationships of a given system [16].

An adaptive network refers a multi-layer feed-forward type structure with interconnect nodes. However, some of the nodes are adaptive, meaning that such a node's output is dependent on several parameters belonging to it [16]. The links in an adaptive network only indicates the flow of information. An adaptive network utilizes a supervised learning algorithm in order to minimize the error of the input/output mapping required, by adjusting the parameters of the adaptive nodes [16]. Therefore, a training dataset is necessary as the training process is similar to that used by neural networks except the parameters of the adaptive nodes are being adjusted instead of the weights of the links in the network.

An ANFIS is a type of adaptive network with the adaptive nodes representing membership functions and the consequent equations along with their corresponding parameters [1]. The goal of an ANFIS is to adjust the membership functions' and consequent equations' parameters to emulate the input/output relationships of a given dataset [21]. Therefore, an ANFIS is functionally equivalent to an FIS except that it has the ability to learn and adapt through a training process using input-output data pairs to discover the most approximate parameters of the FIS to model the system accurately. The basic architecture of a first-order Sugeno (Takagi-Sugeno-Kang) ANFIS with 2 inputs and 2 rules is shown in Figure 5 [1, 16].

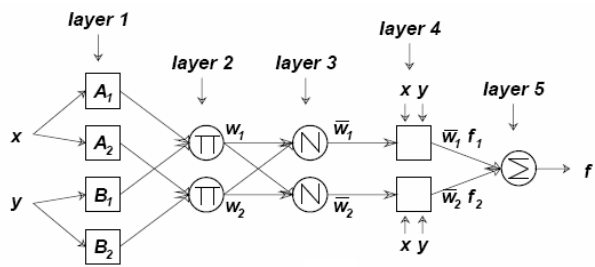

*Figure 5:  Showing the Architecture of First-Order Sugeno ANFIS [1, 16]*

In Figure 5, Layer 1 contains a series of membership functions which determine the degree that the given input belongs to the specific fuzzy set [16]. The membership function's parameters are changed, therefore, changing the shape of the function and the degree of membership of the input to a specific fuzzy set. The nodes in this layer are adaptive and the parameters are known as premise parameters [1].

In Layer 2, each node produces the product of the incoming signals. Thus, determining the final value or firing strength of each rule [16]. The nodes in Layer 2 are fixed, normally performing a fuzzy AND operation.

Each node in Layer 3 calculates the normalized firing strength by taking a rule's firing strengthens and dividing it by the sum of all the rules' firing strengths [16]. Layer 4 is an adaptive layer that has a node function equal to the normalized firing strength multiplied by the first-order Segeno fuzzy model function. The final layer (layer 5) calculates the final output by summing all the incoming signals [16]. Since the normalized firing strength and first-order function are the incoming signals from the previous layer, the output of layer 5 is effectively a weighted average. All the respective equations can be found in [1].

An ANFIS is trained using either back-propagation, or a hybrid training algorithm (a combination of least squares and back-propagation).

## VI. IMPLEMENTATION AND RESULTS

Each of the Artificial Intelligence methods discussed above were implemented using MATLAB. In this section, the implementation, results and observations for each of these methods will be discussed. Also, the pre-processing performed on the Steam Generator dataset is discussed

### A. Data Pre-processing

Before a neural network, SVM, and Neuro-Fuzzy System should be implemented, the dataset needs to be analysed and processed to insure the best possible chance of acquiring the input-output relationship of the dataset. Pre-processing the dataset that will be fed into the neural network or AI system is very important to the performance, generalization ability, and the speed of training of the neural network [3].

On inspection of the dataset, it was seen that there were no data points with missing values but there were a number of outliners. An outliner is an extreme point that does not seem to belong to the dataset and may have an unjustified influence on the model [22]. Since two of the inputs were already scaled between 0 and 1, any samples that had a value greater than 1 or less than zero for either of these scaled inputs were considered to be outliners. The outliners were simply removed from the dataset. There were 965 outliners in the dataset.

Scaling of the data is important in neural networks and SVMs to equalize the importance of each variable [22]. Since different variables can have values that differ in orders of magnitude, the variables with the larger values will appear more significant in determining the outputs [3]. Thus, all inputs should be scaled to have the same range. Also, scaling is important as the activation functions in neural networks only have a limited range before saturation occurs. Both the inputs and the outputs were scaled between 0 and 1 using Min-Max normalization to allow each variable to have equal importance. Min-Max normalization uses the maximum and minimum value of the variable to scale it to a range between 0 and 1, and is given by Equation 28 [22]. The outputs can be converted back to the original scale without any loss of accuracy.

$$x_{Scaled} = \frac{x - x_{min}}{x_{max} - x_{min}}$$

(28)

During the collection of the data, the samples can be stored in a specific order. The dataset stored the samples in the sequential order in which they were captured. Therefore, the samples were randomized in order to break this specific order. Since little else was known about the data, no other pre-process procedures were performed on the data points.

The next step in the pre-processing was to partition the dataset into 3 datasets: training, validation, and a testing dataset. Each dataset should contain a full representation of the available values. The training dataset is used during the supervised training process to adjust the weights and biases to minimize the error between the network's outputs and the target outputs as well as for the training of the SVM and neuro-fuzzy system to adjust their corresponding parameters. The validation data is used to periodically check the generalization ability of the network, SVM, or neuro-fuzzy system. The validation dataset is effectively part of the training process as it is used to guide the selection of the AI system. The test dataset is used as a final measure to see how the AI system performs on unseen data, and should only be used once. The resulting dataset has 8635 records of input-output sets which were divided into the 3 datasets mentioned above. The same 3 datasets were used for the implementation of the neural networks, the SVMs, and the ANFIS.

### B. Performance Measures

The main performance measure that was utilised to evaluate the prediction ability of the Artificial Intelligence Methods was the Mean Squared Error (MSE). The Mean Squared Error is given by Equation 29. This equation allows the contribution of each output to the total MSE to be calculated.

$$MSE = \frac{1}{R} \sum_{k=1}^{R} \|t(k) - y(k)\|^2$$
$$= \frac{1}{R} \sum_{k=1}^{R} \sum_{p=1}^{m} (t_p(k) - y_p(k))^2$$
$$= \sum_{p=1}^{m} \left( \frac{1}{R} \sum_{k=1}^{R} (t_p(k) - y_p(k))^2 \right)$$

(29)

where  *R = size dataset*
       *m = number of outputs*

*y* = predicted value
*t* = desired target value

Other performance measures that were considered are: the time taken to train the AI system, the time taken to execute the AI system, and the complexity of the model produced by the AI method.

*C. Neural Networks Using Standard Approaches*

The neural networks were implemented using the open source NETLAB Toolbox by Ian Nabney. Both the MLP and RBF neural networks were implemented using the standard approaches with this toolbox. The toolbox only constructs a 2 layer feed-forward network for both the MLP and RBF. Therefore, there is only one hidden layer, and only the number of hidden nodes needed to be determined.

The initialisation of the MLP and RBF networks involves determining the activation functions used and the size of the hidden layer. A linear output activation function is best for regression problems, therefore, it was utilised for both the MLP and RBF. In the case of an MLP, a linear output activation function does not saturate, and as a result can extrapolate a little beyond the training dataset [2]. However, the hidden nodes can saturate which is one reason the inputs and outputs were scaled. In NETLAB, the hidden nodes of the MLP are the hyperbolic tangent. The hidden nodes of the RBF used the Gaussian function as seen in Equation 4.

The generalisation ability of a network is determined mainly by the correct model complexity and the number of training cycles. There are a number of methods to improve the generalisation ability of a network such as determining the model complexity, early stopping, and regularisation.

- Model complexity is represented by the number of hidden nodes in the network, as the hidden nodes are responsible for the number of adjustable parameters available in the network [3]. Therefore, a more complex model has a greater number of hidden nodes. However, if there are too many hidden nodes the system will be unnecessarily complex and prone to modelling the system's data too well (over-fitted). Conversely if there are too few hidden nodes the network will not be able to adequately model the system (under-fitted). One way to determine the optimal number of hidden nodes is to train the neural network with different numbers of hidden nodes, and observe the training and validation errors obtained. Note that a large number of hidden nodes will slow the training process.

- The Early Stopping technique uses a training as well as a validation dataset. The main idea behind early stopping is that the training error of the network will gradually decrease as the number of training cycles increases. The degree the network is over-trained is measured using the validation dataset as the validation error will decrease at first and then start to increase as the network is over-trained [3]. Therefore, training should be stopped at the point before the validation error begins to increase.

- Regularisation techniques encourage weights that produce smoother network mappings. An over-fitted network models the training data almost exactly, resulting in the mapping produced by the network having areas of large curvature [3]. This results in large weights. The simplest regularisation technique uses a weight-decay where mappings with large weights are penalised [3]. Regularisation prevents over-training.

The number of inputs and outputs are determined by the problem, and as stated there are 4 inputs and 4 outputs in the system being modelled. Determining the number of hidden nodes used is an iterative and experimental procedure, as it is dependent on the complexity of the relationships in the dataset. A rough estimate for the number of hidden nodes is to take half the sum of the total number of inputs and outputs [2]. Therefore, 4 hidden nodes were used as a starting point and progressively increased while monitoring the training process, to determine an approximate number of hidden nodes. An alternative approach would be to start with a network with a large number of hidden units and prune it iteratively to find a network which will adequately and accurately model the data.

The approach that was taken was to train a network with a fixed number of hidden nodes, periodically stopping the training process to determine the error on the validation dataset. Therefore, the training and validation errors during the training process could be observed, and an indication of the generalisation ability of the network determined. This was done for a varying number of hidden nodes (4 – 20 for the MLP and 4 -50 for RBF). For each number of hidden nodes, a number of networks were trained, as the optimisation techniques used will result in a different solution each time. Therefore, the different solutions for the set number of hidden nodes could be compared and the best network selected. Also, a number of networks with large numbers of hidden nodes were trained to see what the effect was on the resulting network (60, 70 and 200). This procedure was done, in order to determine an optimum number of hidden nodes and training cycles that could adequately and accurately model the system. The goal of this procedure was to find a network that was powerful enough to adequately model the system and generalise well, while being easily trained.

Using the procedure discussed above, it was found that for the MLP the most appropriate number of hidden nodes and training cycles were 8 and 240 respectively. During the experiments carried out on the MLPs, a few observations were made and are discussed below.

It was noticed that increasing the number of hidden nodes beyond 8 didn't seem to increase the accuracy by a significant amount to justify utilising a more complex network. The accuracy obtained for 8 and above hidden nodes was relatively constant, and only slowly increased. Therefore, the least complex network (least number of hidden nodes) with adequate accuracy was chosen as a more network complex will take longer to train and execute. An MLP with 8 hidden nodes has 76 free parameters for the weights and biases. Therefore, increasing the number of hidden nodes increases the number of free parameters that need to be adjusted during the optimisation process.

Normally, the optimum number of training cycles occur at the point were the validation error and the training error start to diverge (Early Stopping). However, since after a certain point the validation and training error remained relatively constant (ran parallel to each other, only slowly decreasing), the point where it started to remain constant was taken as the optimum number of training cycles. This pattern was observed for each MLP network implemented and evaluated, and indicates that the validation and training data must be similar. Refer to Figure 6, showing the pattern observed for 8 hidden nodes for one of neural networks trained. This figure was obtained by periodically stopping the training and noting the validation and training errors.

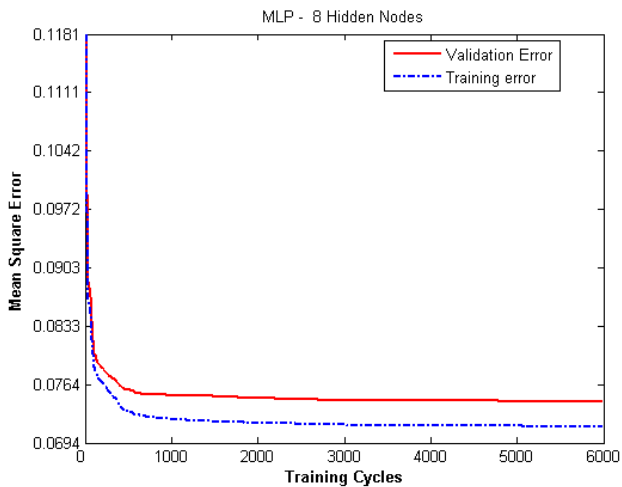

*Figure 6: Showing the MSE vs. the Training Cycles for one of the MLPs Trained.*

For RBF, 30 hidden nodes and 150 training cycles was determined to provide adequate accuracy, comparable to the MLP network implemented. Initially, the number of hidden nodes investigated ranged from 4 to around 20; however, the error obtained was much larger than that obtained for the MLP with the same number hidden of nodes. Therefore, in order to achieve a comparable accuracy to the MLP the number of hidden nodes had to be increased. The best accuracy was obtained with 50 hidden nodes and 100 training cycles. However, once again increasing the number of hidden nodes above 50 resulted in the training error decreasing substantially but the validation error remained relatively constant. Therefore, it was decided that 50 hidden nodes were appropriate as beyond this there was not a great deal of improvement on the validation error. Also, the accuracy difference between that of 50 and 30 hidden nodes was comparatively small. Determining the number of training cycle necessary for RBF was not as easy as it was for the MLP, as the validation and training error was more "jumpy" than was observed with the MLP. However, the validation error was relatively steady after a certain point and did not increase: 150 for 30 hidden nodes and 100 for 50 hidden nodes.

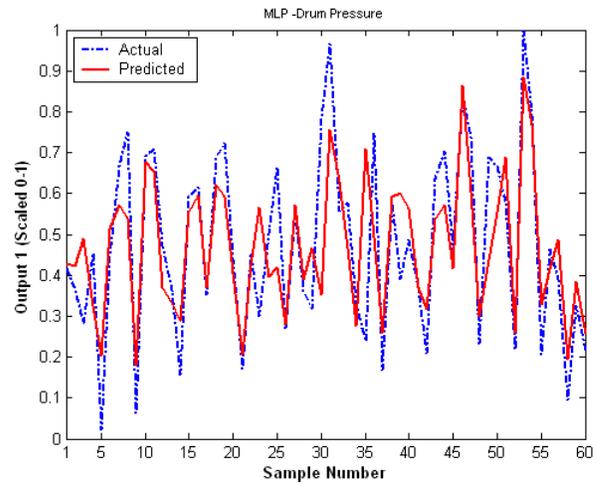

*Figure 7: Showing the Predicted vs. Actual Values for the first 60 points of Output 1 for the Test Dataset applied to the MLP*

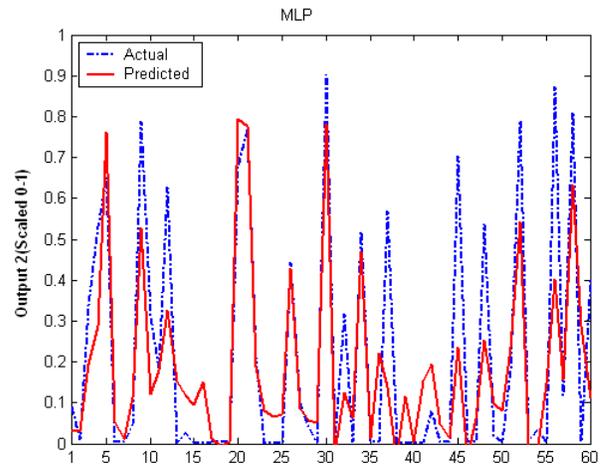

*Figure 8: Showing the Predicted vs. Actual Values for the first 60 points of Output 2 for the Test Dataset applied to the MLP*

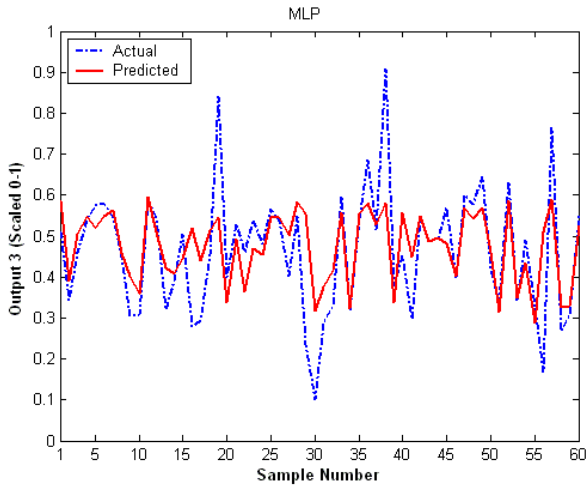

*Figure 9: Showing the Predicted vs. Actual Values for the first 60 points of Output 3 for the Test Dataset applied to the MLP*

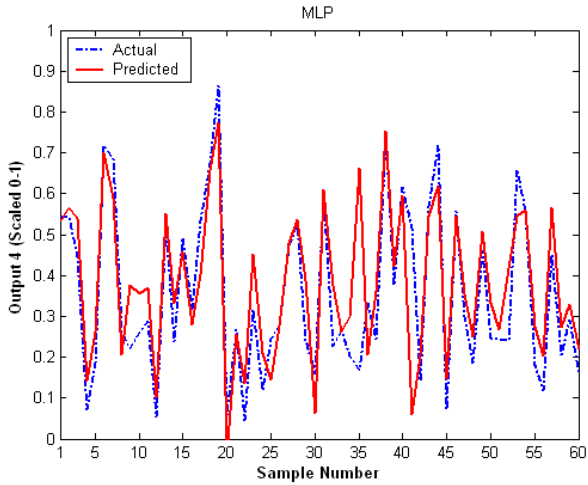

*Figure 10: Showing the Predicted vs. Actual Values for the first 60 points of Output 4 for the Test Dataset applied to the MLP*

A problem encountered with the RBF network implementation was that the training function for the RBF network in NETLAB (rbftrain) encounters a divide by zero error when the number of hidden nodes was substantially large and the code had to be modified if simulations were run with large number of hidden nodes. Alternatively, the training function for the MLP (netopt) could have been used for the RBF training; however, it no longer uses the 2 stage training process.

A combination of early stopping and regularisation was used to determine theses optimum parameters. The values for alpha (weight decay coefficient) and beta (Inverse Noise ratio) were initially set to the default values. However, they didn't have to be changed significantly and were eventually set to 0.01 and 1 respectively.

It was noticed that by adjusting either the hidden nodes or the number of training cycles, that different outputs contributed more to the overall error of the system (Equation 29). As a result, if an attempt was made to improve the network accuracy with respect to a certain output, it was found that the accuracy decreases with respect to one or more of the other outputs. The MLP is effectively trying to model 4 separate functions at once, therefore, the hidden nodes may have been having difficulty learning in order to model all 4 functions at the same time. This is referred to as cross-talk [2]. One way to attempt to solve this problem would be to model each output as a separate network [2].

For both the MLP and RBF, it was difficult to model Output 3. This can be seen from the Actual vs. Predicted plots of the first 60 data points for the testing dataset in Figures 9 and 13 for Output 3. An attempt was made to decrease the error contribution of Output 3 by adjusting the number of training cycles and hidden nodes. However, it made a small difference to the error that Output 3 contributed to the total error, and caused the contribution to the total error of the other outputs to increase. The neural networks did not seem to be able to model Output 3 as accurately as the other outputs of the system. A possible reason maybe that the dependency of Output 3 on the given inputs is weak, therefore, more input variables may need to be measured in order to model this output more accurately.

The following performance measures were evaluated for each of the neural networks implemented: (i) the time taken to train the network using the training dataset, (ii) the time taken to execute or forward-propagate through the network for the testing dataset and (iii) the MSE accuracy obtained by the network on the testing dataset. The results are summarised in Table 1, for the optimum networks obtained for the RBF and MLP. The scaled conjugate gradient algorithm is used to optimise the MLP weights and biases. The RBF network with 30 hidden nodes is shown below as it has a comparable accuracy to the MLP obtained.

*Table 1: Performance Characteristics for Individual MLP and RBF Networks*

|  | MLP | RBF |
|---|---|---|
| **Time to Train (s)** | 6.98 | 13 |
| **Time to Execute (s)** | 0.016 | 0.031 |
| **MSE of Test Dataset** | 0.075708 | 0.076360 |
| **No. of Hidden nodes** | 8 | 30 |
| **No. of Training Cycles** | 240 | 150 |

From Table 1, it can be seen that the RBF with comparable accuracy to the MLP took much longer to train. The complexity of the MLP and RBF with comparable accuracy is significantly different. The MLP has 8 hidden nodes corresponding to 76 free parameters while the RBF has 30 hidden nodes corresponding to 274 free parameters. While the RBF is supposed to be faster during the training process [2], the increased complexity of the network has increased the training time significantly. Both the MLP and RBF gave similar accuracy. The MLP was faster to execute than the RBF, which was expected. The plots for the Actual vs. Predicted values for the first 60 points for each output for the MLP are shown in Figures 7-10, and for the RBF in Figures 11-14.

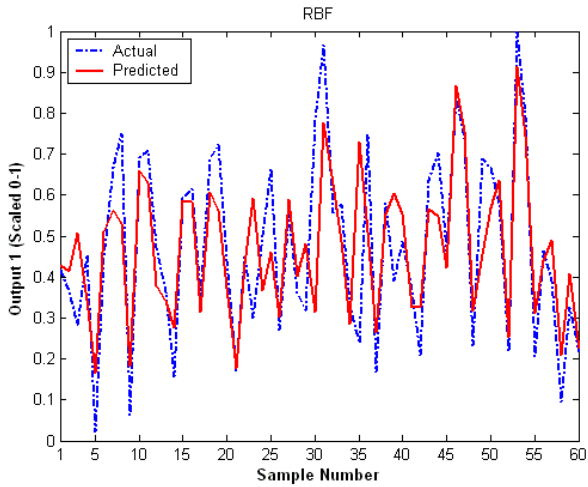

*Figure 11: Showing the Predicted vs. Actual Values for the first 60 points of Output 1 for the Test Dataset applied to the RBF*

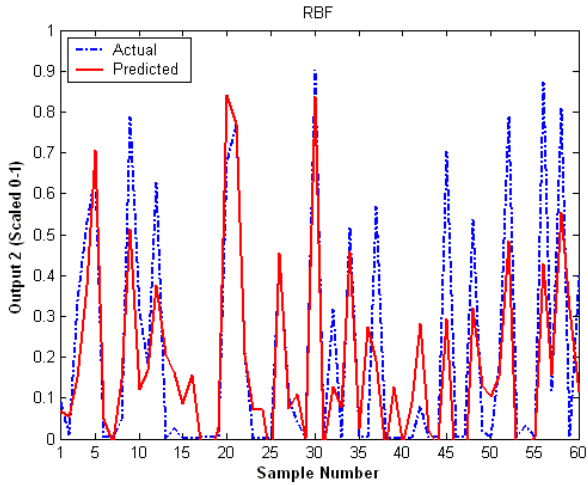

*Figure 12: Showing the Predicted vs. Actual Values for the first 60 points of Output 2 for the Test Dataset applied to the RBF*

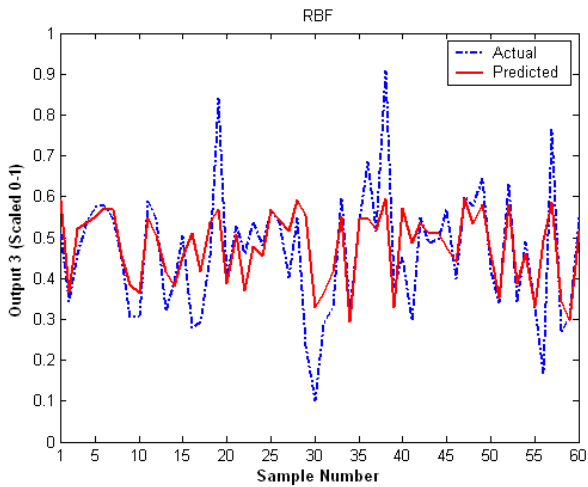

*Figure 13: Showing the Predicted vs. Actual Values for the first 60 points of Output 3 for the Test Dataset applied to the RBF*

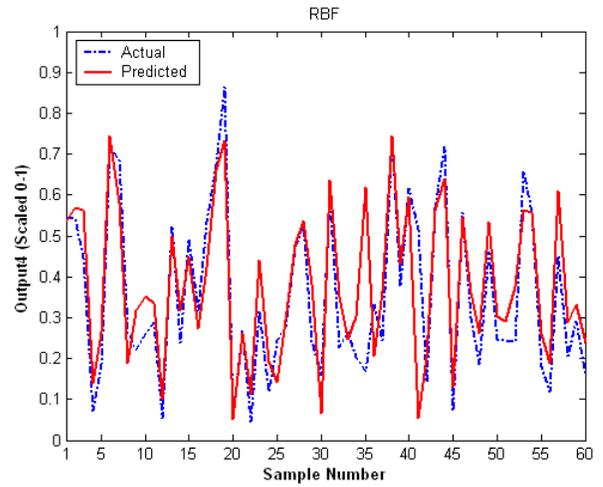

*Figure 14: Showing the Predicted vs. Actual Values for the first 60 points of Output4 for the Test Dataset applied to the RBF*

An RBF network with the same number of hidden nodes as the MLP was implemented to observe how long it would take to train and execute in comparison to the MLP. If the networks have similar complexities it was observed that the RBF trained faster than the MLP, and forward executed with a similar speed to the MLP. The RBF network with 8 hidden nodes that was implemented took 3 seconds to train and 0.016 seconds to execute. However, the RBF model with the 8 hidden nodes had a larger MSE than the MLP.

### D. Committees

A simple averaging committee was implemented for both the MLP and RBF using the NETLAB Toolbox. The averaging committee consisted of 10 MLP networks with identical architectures. Each network in the committee was trained using the same training dataset, however, each network was initialised differently and trained independently. The final output was taken as the average of the individual outputs for each network. Also, a committee consisting of 10 RBF networks was constructed the same way as described for the MLP committee. The architectures of each network in the committees used the optimum parameters found using the standard approaches.

It was found that the averaging committees only marginally improved the accuracy obtained. The committees took longer to train and execute which can only be expected as the committee is effectively 10 networks. Table 2, shows the results captured for the committee implemented. From Equation 5, increasing the number of neural networks in the committees would result in the error reducing as the number of neural networks increased.

*Table 2: Showing the results for the committee networks consisting of neural networks with identical architectures.*

|  | MLP Committee | RBF Committee |
|---|---|---|
| **MSE of Test Dataset** | 0.0746639 | 0.076200 |
| **Time to Train (s)** | 80 | 129 |
| **Time to Execute (s)** | 0.172 | 0.407 |

Another implementation of a committee that was tested was bagging. In bagging, each network is trained on a bootstrap dataset. A bootstrap dataset is a dataset that is randomly created by selecting *n* points with replacement, from the training dataset with *n* patterns [23]. This means that some data points are chosen more than once and are duplicated in the bootstrap dataset, while some data points will not be selected at all. Then each bootstrap dataset created is used to train a separate network, and the final output of the committee is calculated by averaging the outputs of the networks created [23].

A committee of 10 neural networks was created and each was trained with a bootstrap dataset. The training time was slightly longer than that of the straight averaging committee as the training time included the time taken to create the bootstrap datasets. The committee created using bagging increased the accuracy slightly from that of the simple averaging network but not significantly. The results obtained for the committee using bagging are in Table 3.

*E. Bayesian Techniques for Neural Networks*

The architectures of the MLP and RBF used for the Bayesian techniques were the optimum architectures (number of hidden nodes, number of inputs and outputs, activation functions) found using the standard approaches discussed in the previous sections. This allows comparisons to be made between the results obtained from both approaches.

*Table 3: Showing the results for the committee networks using bagging*

|  | MLP Committee (Bagging) | RBF Committee (Bagging) |
|---|---|---|
| **MSE of Test Dataset** | 0.074455 | 0.075202 |
| **Time to Train (s)** | 90 | 144 |
| **Time to Execute (s)** | 0.172 | 0.5 |

The Bayesian techniques were implemented using NETLAB for the neural networks. NETLAB allows the implementation of the Bayesian techniques to be done using the Hybrid Monte Carlo algorithm. The Bayesian Network utilizing Hybrid Monte Carlo algorithm is implemented using NETLAB by the following steps: the sampling is executed, each set of sampled weights obtained are placed into the network in order to make a prediction, and then the average prediction is computed from the predicted values obtained from each set of sampled weights [3]. Since Bayesian Techniques don't require cross-validation techniques to determine parameters, a larger training dataset could be used.

For the Hybrid Monte Carlo algorithm the following parameters were adjusted to determine the best set of parameters to model the dataset: the step size, the number of steps in each Hybrid Monte Carlo trajectory, the number of initial states that were discarded, and the number of samples retained to form the posterior distribution.

At first a step size of 0.005 was chosen, however, this resulted in a large number of the samples being rejected. Therefore, step sizes less than 0.005 were utilised and the results from the experiments noted. It was found that any step size above 0.001 had high rejection rate, and therefore, a low acceptance rate. As a result, step sizes of 0.001 and 0.0005 were tested along with the other parameters. A step size of 0.0005 gave a 96% acceptance rate and the results are shown in Tables 4 and 5. If the step size is extremely small, the Hybrid Monte Carlo algorithm will take a long time to converge to a stationary distribution as the state space is being explored in much smaller steps. If the step size is large then too much exploration may occur causing the Hybrid Monte Carlo algorithm to "jump" over the distribution that is being searched for, effectively missing it.

It was noticed that increasing the number of samples retained did not improve the accuracy of the network. Therefore, 100 samples were eventually retained which is a relatively small number of samples. The number of steps in a trajectory were modify for different runs, however, after a certain point it didn't seem to improve the accuracy and the steps in a trajectory were set to 100. It was observed that too few steps in a trajectory didn't allow enough of the sample space to be explored and the MSE was larger for a smaller number of steps. The number of samples omitted was chosen by observing the average number of samples rejected at first, since, once the other parameters had been chosen the acceptance rate was high, the number of samples omitted was set to a reasonably small number of 10. The coefficient of data error (β) was varied and eventually set to 30. The coefficient of weight-decay prior was set to the same used for the standard approach. Tables 4 and 5, show the results obtained for some of the networks implemented for the Bayesian MLP. In Table 5, Network 6 gives the best results, and the measures for this network are shown in Table 6.

*Table 4: Showing some of the results obtained in the implementation of the MLP using Bayesian techniques (Hybrid Monte Carlo)*

|  | Network 1 | Network 2 | Network 3 |
|---|---|---|---|
| **MSE for Testing Dataset** | 0.078884 | 0.0742217 | 0.0746754 |
| **MSE for Training Dataset** | 0.079786 | 0.072708 | 0.073635 |
| **Step Size** | 0.001 | 0.001 | 0.001 |
| **No. of Samples Retained** | 100 | 100 | 100 |
| **No. Initial States Omitted** | 10 | 10 | 10 |
| **No. of Steps in a Trajectory** | 100 | 100 | 200 |
| **β** | 1 | 30 | 30 |
| **α** | 0.01 | 0.01 | 0.01 |

*Table 5: Showing some of the results obtained in the implementation of the MLP using Bayesian techniques (Hybrid Monte Carlo)*

|  | Network 4 | Network 5 | Network 6 |
|---|---|---|---|
| **MSE for Testing dataset** | 0.075889 | 0.075575 | 0.073714 |
| **MSE for Training dataset** | 0.076826 | 0.074250 | 0.072730 |
| **Step Size** | 0.001 | 0.001 | 0.0005 |
| **No. of Samples Retained** | 200 | 200 | 100 |
| **No. Initial States Omitted** | 10 | 10 | 10 |
| **No. of Steps in a Trajectory** | 200 | 100 | 100 |
| **β** | 30 | 30 | 30 |
| **α** | 0.01 | 0.01 | 0.01 |

*Table 6: Showing the performance measures taken for Network 6*

|  | Bayesian MLP |
|---|---|
| **Training Time (s)** | 212.8 |
| **Execution time (s)** | 1.4 |

From, the results in Tables 4 - 6, it can be seen that the Bayesian MLP gave a better accuracy than the single MLP implemented using standard approaches. However, it took a substantial amount more time to train and execute compared to the single MLP.

The Bayesian techniques using Hybrid Monte Carlo were attempted with an RBF, however, difficulties were experienced and no definite results were obtained.

*F. Support Vector Machines*

The LS-SVMlab Toolbox for Matlab was used to simulate the SVM for the given dataset. This toolbox implements Least Squares Support Vector Machines for both classification and Regression problems [24]. Another toolbox that implemented the $\varepsilon$-insensitivity loss function SVM was found, however, to run a simulation was extremely time consuming even when the number of samples used to train the SVM were substantially decreased.

Since an SVM determines an unknown dependency between a set of inputs and an output, the toolbox handles the case of multiple outputs by treating each of the outputs separately. Therefore, there are effectively 4 SVMs modelling the dataset. This is different to the neural networks where one network was trained to model all 4 outputs. As a result, 4 SVMs were simultaneously implemented and trained, one for each output of the dataset.

The implementation of the LS-SVM required the selection of two free parameters since a Radial Basis function was used for the kernel function. Therefore, the optimum values of the two free parameters needed to be determined: the width or bandwidth of basis function ($\sigma^2$), and the regularisation or penalty parameter (C). An empirical approach was taken in determining the 2 free parameters, and is similar to the approach taken in [13]. Since there are 4 outputs, the parameters for each corresponding SVM had to be determined. The procedure used is discussed below with respect to the determination of the parameters for modelling Output 1. The same procedure was used for the determination of the parameters for the other outputs in the dataset.

First, the regularisation constant was set at a value of 10 while varying the bandwidth of the basis function for training data corresponding to Output 1. The basis function's width was varied for values from 0.3 to 1000. For a small $\sigma^2$, the training error was at its minimum; however, the validation error was very large. This gives an indication that the LS-SVM is over-trained for small $\sigma^2$. At around $\sigma^2 = 1$, the training and validation errors crossed and remained constant for a while. Then from about $\sigma^2 = 10$, both the validation and training error started to increase which indicates that the SVM was not even able to model the training data well for large values of $\sigma^2$, and is under-trained. An appropriate choice for the bandwidth of the basis function was decided to be 1, from the above experiments carried out.

Secondly, the bandwidth of the basis function was kept constant at 1 while the value of the regularisation constant was varied. The regularisation constant was varied between 1 and 1000 while observing the training and validation errors. As the regularisation constant (C) was increased both the validation and training error decreased together, until a certain point where the validation error started to increase while the testing error continued to decrease. Thus, for a small value of C it appears to under-fit the training data, and for large values of C the SVM appears to over-fit the training dataset. Therefore, the most appropriate value for the regularisation constant was 10, as beyond this value the validation error starts to increase. The optimum parameter values chosen to model Output 1 were C=10 and $\sigma^2= 1$. The optimum parameters for each of the SVM corresponding to the 4 outputs can be seen in Table 7. Figures 14-17 show the Actual vs. the Predicted values of the first 60 samples of the each output for test dataset applied to the LS-SVMs.

*Table 7: Showing the results obtained for the implementations of the LS-SVM*

| SVM | Output1 | Output2 | Output3 | Output4 |
|---|---|---|---|---|
| **MSE for Test Dataset** | 0.023300 | 0.023600 | 0.01125 | 0.015400 |
| **Training Time (s)** | 90 | 60 | 50 | 120 |
| **Execution Time (s)** | 2.7 | 2s | 2.7 | 2.6 |
| $\sigma^2$ | 1 | 1 | 10 | 0.1 |
| C | 10 | 1 | 10 | 10 |

From Table 7, it can be seen that the LS-SVM took longer to train and execute than the neural networks produced using the standard approach. Even tough, the neural network was modelling 4 relationships it was much faster than the LS-SVM which is only modelling one relationship at a time. The results obtained from LS-LVM were easily reproducible as opposed to neural networks where one can easily obtain different and less accurate results when the simulation is rerun, due to the optimisation techniques used.

If the error of each of the SVMs are added together as if they are working in a committee to predicted each output of the Steam Generator separately, the effective MSE would be approximately 0.07355.

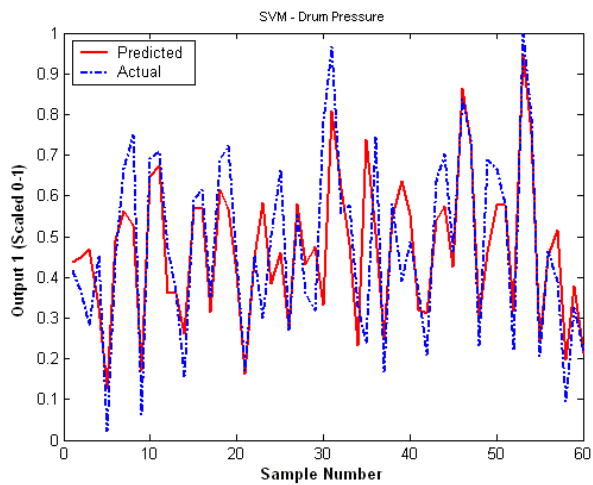

*Figure 14: Showing the Predicted vs. Actual Values for the first 60 points of Output 2 for the Test Dataset applied to the LS-SVM*

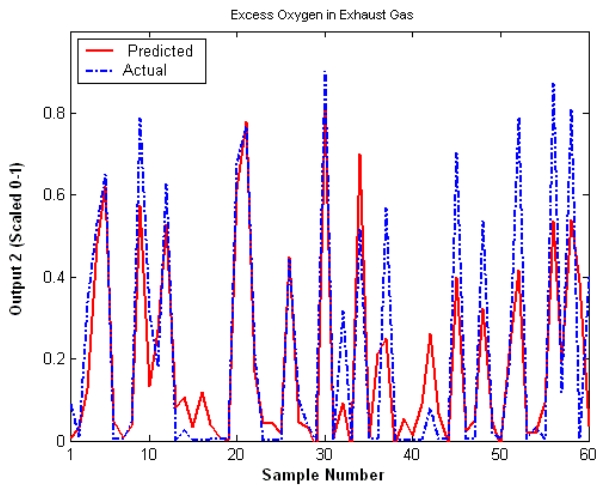

*Figure 15: Showing the Predicted vs. Actual Values for the first 60 points of Output 1 for the Test Dataset applied to the LS-SVM*

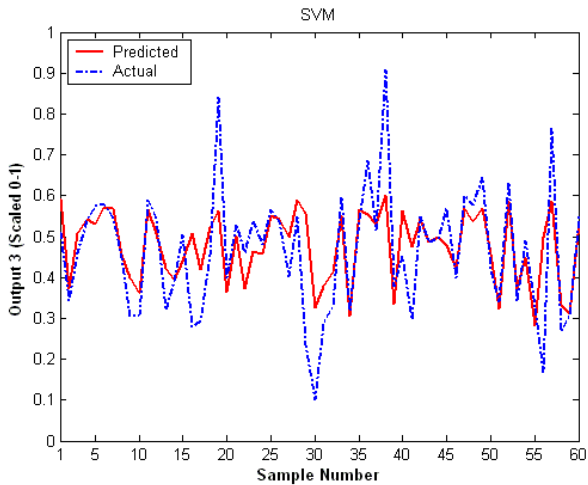

Figure 16: Showing the Predicted vs Actual Values for the first 60 points of Output 3 for the Test Dataset applied to the LS-SVM

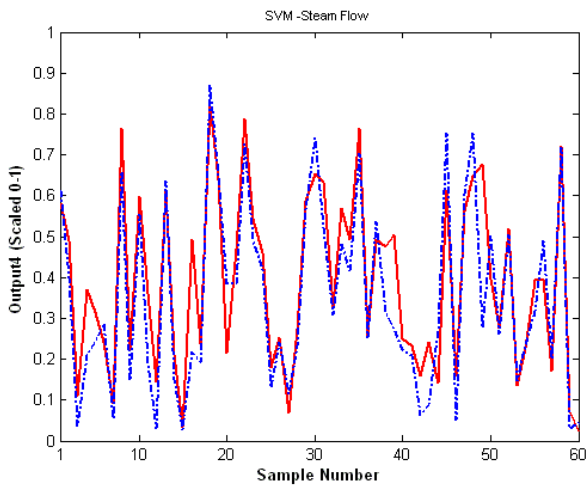

*Figure 17: Showing the Predicted vs. Actual Values for the first 60 points of Output 4 for the Test Dataset applied to the LS-SVM*

## G. Adaptive Neuro-Fuzzy Systems

The Fuzzy Logic Toolbox for Matlab was used to simulate the Adaptive Neuro-Fuzzy System. The training process involves modifying the membership function parameters of the FIS in order emulate the training dataset to within some error criteria [21]. The toolbox implements a Sugeno-type system for the Adaptive Neuro-Fuzzy Inference System. It only supports a single output which is obtained using a weighted average defuzzification process. In the Fuzzy Logic Toolkit, the number of output membership functions must be equal to the number of rules generated, and the output membership functions must be linear or a constant. Therefore, each output for the given dataset was modelled separately, and a linear output function of the form in Equation 27 was used.

The toolkit allows for a training and validation dataset to be used, in this case the Toolbox selects the model with the minimum validation data error [21]. The idea is that overtraining will be avoided, as it is expected that the validation error will decrease as training takes place until a certain point where the validation error begins to increase, indicating overtraining. The learning process is similar to neural networks except that different parameters are being adjusted.

The idea is to tailor the membership functions to model the input/output relationship of the dataset [21]. The ANFIS constructs a FIS in which its membership function parameters adjusted by a training algorithm. In this way, the parameters of the membership functions will change through the process of learning. The Toolkit uses either back-propagation or a hybrid method (least squares and back-propagation) to train the ANFIS.

The Fuzzy Logic Toolbox has 11 different membership functions available, of which 8 can be used with the Adaptive Neuro-Fuzzy System: Triangular function, trapezoidal, 2 different Gaussian functions, bell function, Sigmoidal Difference function (difference of 2 Sigmoidal functions), Sigmoidal product function (product of 2 Sigmoidal functions), and polynomial Pi curves. The Sigmoidal functions have the property of being asymmetrical opposed to the Gaussian which is symmetrical in nature [21].

In the Fuzzy Logic Toolkit the number of input membership functions and the type of membership function used could be modified. The number of membership functions was left at the default of 2 per input, giving 8 input membership functions. Firstly, an FIS structure was initialised which could then be adjusted to model the dataset provided. The generated FIS structure contained 16 fuzzy rules, therefore, 16 output membership functions. The ANFIS function keeps track of the Root Mean Square Error (RMSE) of the training dataset at each epoch, as well as the validation error associated with a validation dataset.

Different input membership functions were tried for each output being modelled. A curve of the training and validation errors vs. the training cycles was observed by plotting the values stored by the ANFIS function in the toolbox. It was then possible to see how many epochs were necessary and the generalisation ability of the ANFIS produced. A final test dataset was used to verify the generalization ability of the ANFIS; and the actual vs. the predicted values for the first 60 samples of the test dataset for that output was plotted. This process was done for each output.

First, Output 1 was modelled using different Membership functions. It was found that the Gaussian, Sigmoidal, and Polynomial Pi Function could all model the data with fairly reasonable accuracy; and the training time tended to vary according to which membership function was used. The bell and Gaussian Function took a long time to train, while the sigmoid was the fastest to train. However, the Polynomial Pi Membership Function gave the best accuracy overall, and was reasonable fast to train in comparison to ANFIS using the Gaussian and bell Membership functions. Both the Triangular and trapezoidal membership functions appeared to be too simple to model the underlying complexities of the data given.

The plots of the training and validation errors vs. the training cycles for the ANFISs, trained using some of the difference membership functions for Output 1 are shown in Figures 18-21. It can be seen in Figure 20 that using a triangular input membership function, the validation error is erratic at first, jumping up and down, and the error quickly increases. However, the validation error for the ANFIS using the other membership functions rapidly decreases, and then remains relatively constant. This gave an indication that the training and validation dataset must be relatively similar to a degree. Table 8, shows the results obtained for the simulations done for Output 1, and Figure 22 shows the Actual vs. Predicted values for first 60 samples of the test dataset for Output1 using a Polynomial Pi input membership function.

*Table 8: Showing the results obtained for the simulations done for Output 1 for the ANFIS*

| Output 1 | Gaussian Function | Sigmoidal Difference Function | Polynomial Pi Curve |
|---|---|---|---|
| **MSE for Test Dataset** | 0.022608 | 0.022663 | 0.022564 |
| **Training Time (s)** | 440.11 | 39.2 | 55.3 |
| **Execute Time (s)** | 0.047 | 0.047 | 0.047 |
| **No. of Training Cycles** | 400 | 35 | 50 |
| **No. Fuzzy Rules** | 16 | 16 | 16 |

The same procedure was followed to model the input/output relationship for Output 2. For the ANFISs trained using the Sigmoidal and Triangular membership functions, a slight increase in the validation error was observed after a certain number of training cycles. However, the validation error for the ANFIS using the other membership functions rapidly decreases, and then remains relatively constant. The results for the ANFIS for Output 2 are shown in Table 9. The Polynomial Pi Membership function produced the best results, and didn't take too long to train. Figure 23, shows the Actual vs. Predicted values for first 60 samples of the test dataset for Output 2 using a Polynomial Pi membership function.

*Table 9: Showing the results obtained for the simulations done for Output 2 for the ANFIS*

| Output2 | Gaussian Function | Sigmoidal Difference Function | Polynomial Pi Curve | Triangular Function |
|---|---|---|---|---|
| **MSE for Test Dataset** | 0.023436 | 0.02346788 | 0.022253 | 0.023303 |
| **Training Time (s)** | 131.33 | 36.25 | 66 | 191.2 |
| **Execute Time (s)** | 0.032 | 0.031 | 0.046 | 0.062 |
| **No. of Training Cycles** | 120 | 33 | 60 | 175 |
| **No. Fuzzy Rules** | 16 | 16 | 16 | 16 |

The sigmoidal difference function produced the most accurate result for the ANFIS for Output 3. The error has decreased substantially from what was obtained for the SVMs. However, if a plot of the predicted vs. actual is observed (Figure 24), the ANFIS cannot model the values below 0.3 or above 0.6 accurately, but the points between these values seem to be modelled fairly accurately. After looking at the source data for Output 3, it was noticed that most of the data points for Output 3 were between 0.3 and 0.6. The data points above 0.6 and below 0.3 only account for approximately 20% for all 3 datasets. Therefore, the error went down as the majority of the points were being accurately modelled. Some investigation on how Output 3 is actually related to the inputs and possibly some extra pre-processing may be required in order to allow Output 3 to be more accurately modelled. Also, during the training process the validation error increased instead of decreasing at first for certain of the membership functions. If the actual vs. predicted was plotted for the training dataset it looked similar to the test dataset plot, indicating that the input/output relationship for Output 3 is not being modelled adequately by the ANFIS. Table 10 shows the performance measures for Output 3, and Figure 24 shows the Actual vs. Predicted values for first 60 samples of the test dataset for Output 3.

*Table 10: Showing the results obtained for the simulations done for Output 3 for the ANFIS*

| Output 3 | Sigmoidal Difference Function |
|---|---|
| **MSE for Test Dataset** | 0.009436 |
| **Training Time (s)** | 109 |
| **Execute Time (s)** | 0.046 |
| **No. of Training Cycles** | 100 |
| **No. Fuzzy Rules** | 16 |

The Polynomial Pi Membership Function produced the most accurate results for modelling Output 4. The Gaussian membership function was not appropriate this time as the validation error actually only increased and didn't decrease at all. All the ANFISs trained for Output 4 produced exceptionally accurate results, which could be seem from the plots of the predicted vs. the actual. Table 11, shows the performance measures for Output 4, and Figure 25 shows the Actual vs. Predicted values for first 60 samples of the test dataset for Output4 using a Polynomial Pi membership function.

*Table 11: Showing the results obtained for the simulations done for Output 4 for the ANFIS*

| Output 4 | Polynomial Pi Curve |
|---|---|
| **MSE** | 0.014330 |
| **Training Time (s)** | 132 |
| **Execute Time (s)** | 0.046 |
| **No. of Training Cycles** | 120 |
| **No. Fuzzy Rules** | 16 |

The Adaptive Neuro-Fuzzy Inference System was easy to implement and the results obtained show that it can accurately model a system as shown by Output 4. The improvement in the accuracy for Output 4 was significant. The simulations for the ANFIS produced better accuracy than the SVMs and had similar training time. However, the ANFIS executed much faster than the SVMs. Summing the MSE of each ANFIS to produce the effective error of the 4 ANFIS working as a committee to predict the steam generator outputs, gives an approximate MSE of 0.06858.

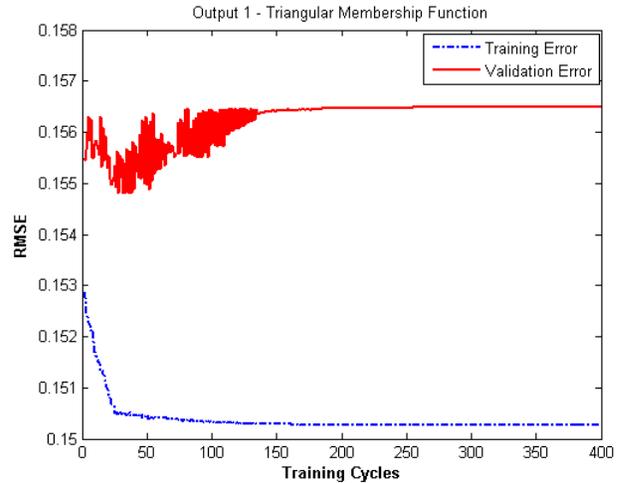

*Figure 19: Showing the Training and. Validation error vs. the training cycles for the ANFIS using Gaussian Input Membership Functions for Output 1.*

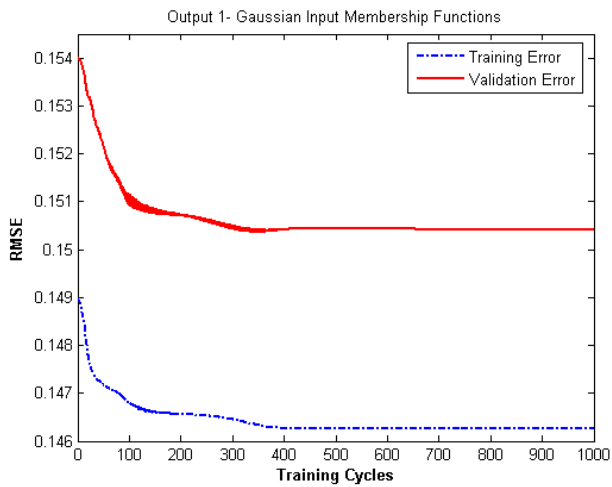

*Figure 18: Showing the Training and. Validation error vs. the training cycles for the ANFIS using Gaussian Input Membership Functions for Output 1.*

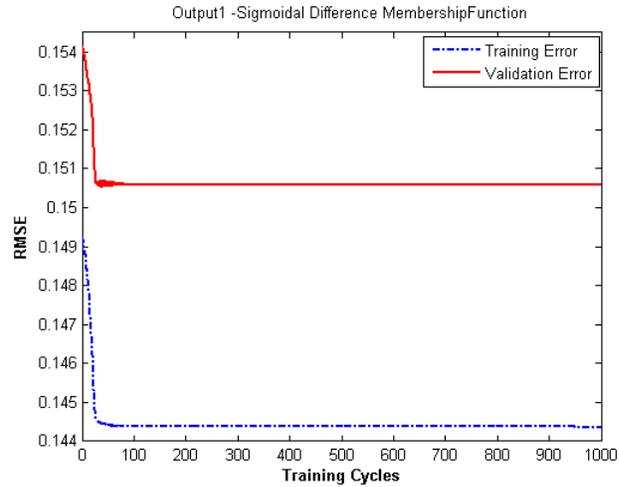

*Figure20: Showing the Training and. Validation error vs. the training cycles for the ANFIS using Gaussian Input Membership Functions for Output 1.*

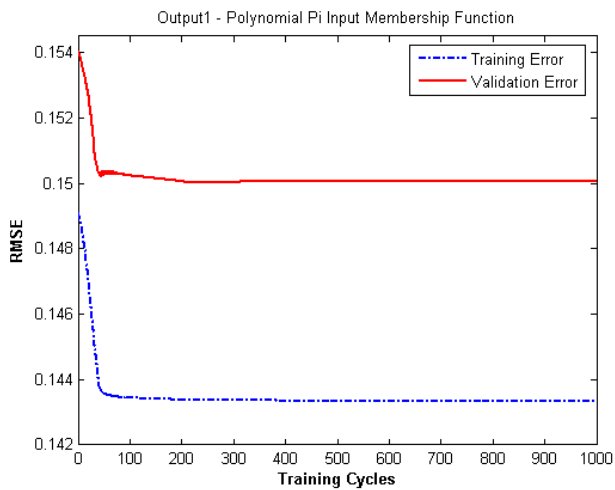

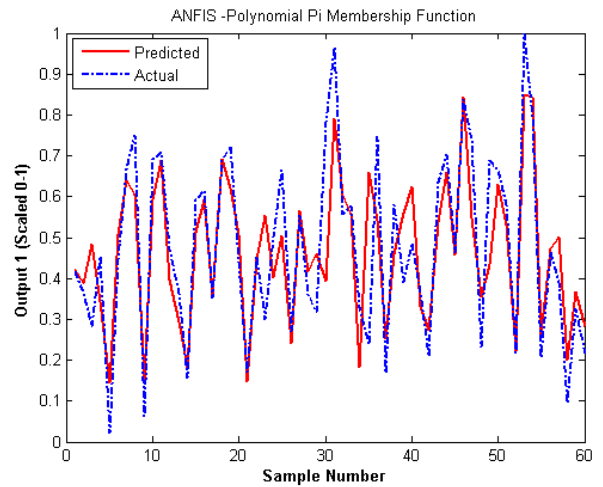

*Figure 21: Showing the Training and. Validation error vs. the training cycles for the ANFIS using Gaussian Input Membership Functions for Output 1.*

*Figure 22: Showing the Predicted vs. Actual Values for the first 60 points of Output 1 for the Test Dataset applied to the ANFIS*

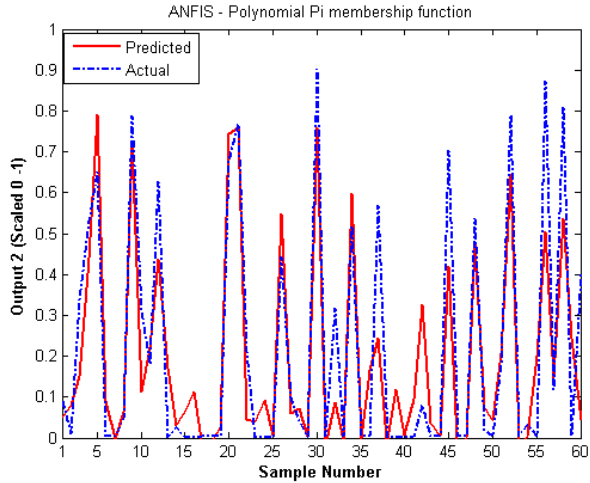

*Figure 23: Showing the Predicted vs. Actual Values for the first 60 points of Output 2 for the Test Dataset applied to the ANFIS*

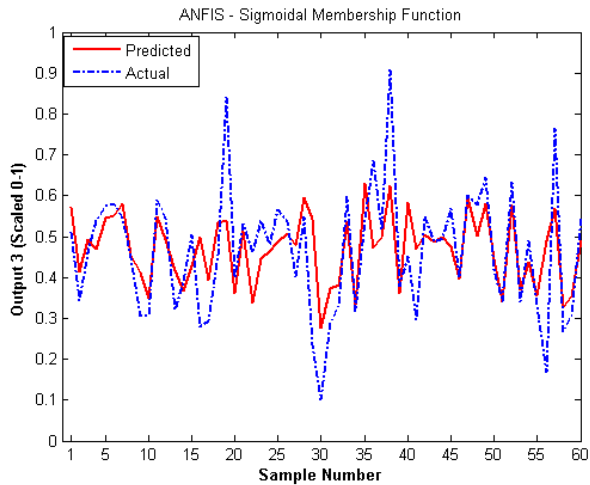

*Figure 24: Showing the Predicted vs. Actual Values for the first 60 points of Output 3 for the Test Dataset applied to the ANFIS*

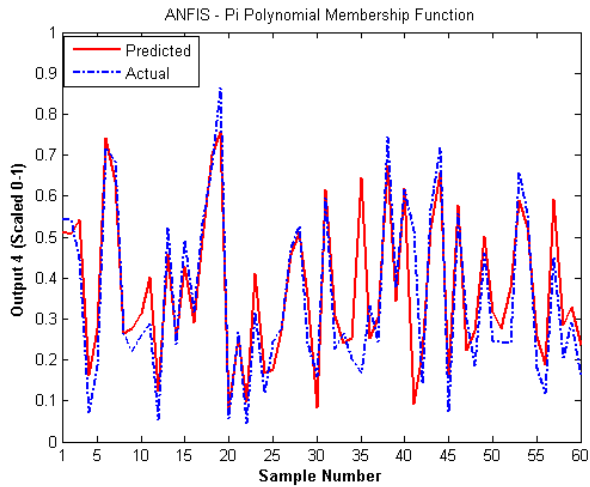

*Figure 25: Showing the Predicted vs. Actual Values for the first 60 points of Output 3 for the Test Dataset applied to the ANFIS*

## VII. DISCUSSION

All the Artificial Intelligence Methods investigated and simulated were capable of generalising on unseen data reasonably well. This section will discuss the principle findings of the investigations and simulations carried out.

The neural networks were difficult to tune as they had the most parameters to adjust. Therefore, finding the optimum parameters to model the given steam generator dataset was a tedious task. Also, the results were not easily reproducible due to the optimisation algorithm used. Therefore, many networks had to be trained in order to achieve accurate results and the best network found was retained.

It was found that the MLP and RBF with similar accuracy had significantly different complexities. In order, to obtain an RBF with a comparable accuracy to that of the MLP produced, the number of hidden nodes in the RBF had to be increased to over 3 times the number of that of the MLP. While the RBF is supposed to be faster during the training process [2], the increased complexity of the RBF network resulted in a significantly increased training time.

The averaging committees of neural networks only slightly increased the accuracy obtained. Since the number of neural networks being trained and executed was more than one, the training and execution time increased compared to the time obtained for an individual neural network. The committee constructed using bagging improved the accuracy slightly from that of the simple averaging committee, however, it took slightly longer to train as the bootstrap dataset had to created.

Using the Bayesian techniques for the MLP improved the accuracy obtained by the MLP. However, it increased the training and execution time required by the neural network produced. This result should be expected as the Bayesian technique is similar to a committee of 100 networks since 100 samples were retained.

The SVM was more accurate than the neural networks. However, a separate SVM had to be implemented for each output. Also, the SVMs had less parameters to tune, thus, making them much easier to implement. The SVM took a long time to train in comparison to the neural networks even though the neural networks were modelling all four relationships between the inputs and outputs at once. Also, the SVMs were slower to execute on unseen data. The SVMs produced were of a comparable accuracy to the Bayesian MLP. Yet, the Bayesian MLP took at least twice the amount of time to train, while the Bayesian MLP was faster to execute. The results obtained from the SVMs were easily reproduced.

The ANFIS out performed all the other methods. The training time for the ANFIS was more than the neural networks but still less than the other methods. However, it should be noted that the training time was highly dependent on the input membership function chosen. Therefore, the membership function with the best accuracy and least training time was selected for each ANFIS implemented. The execution time for the ANFIS was fast and comparable to the execution time obtained for the neural networks.

Also, the ANFISs were easily implemented as there were only two items that could be changed: the number input membership functions per an input variable, and the type of membership function used. Despite the fact that, only 2 input membership functions were used per input, the ANFIS was able to accurately model the outputs, and out perform the other AI methods tested. It would be of benefit to try more membership functions per input to see if the accuracy would improve significantly, however, the training time would increase if this was attempted. From the figure of the predicted vs. the actual for Output 4, it can be seen that this method models the relationship necessary for Output 4 extremely well.

An observation that was made was that certain outputs or relationships were better modelled than others. This may be due to the relationship present in the given steam generator data, therefore, the outputs modelled more accurately may have stronger dependencies on the given inputs. It was seen that Output 3 had a high accuracy but the actual vs. predicted plots showed that certain output points were not being modelled well. There may be several reasons for this result, some of which have been mentioned in the sections above. A further investigation should be done if Output 3 were to be better represented. Also, this may be due to the fact that little is known about the dataset and more pre-processing may be required to eliminate data points that bias the training in some way. Another normalisation method could be tried for the pre-processing calculation.

The optimum parameters selected probably are not the best parameters that could be obtained if an exhaustive search was performed. However, an exhaustive search is computationally expensive and impractical to perform in reality. Therefore, a more empirical approach was used to select the free parameters for each of the AI methods implemented; making it a difficult task to obtain the optimum combination of the parameters which produces the best prediction performance.

## VIII. CONCLUSION

All Artificial Intelligence methods investigated were capable of modelling the steam generator data. The ANFIS out performed the other methods giving the best accuracy overall. It was obvious from the plots of the predicted vs. actual outputs that the methods were able follow the general shape of the actual output data points. The Bayesian neural networks and the SVM gave comparable accuracy. The standard neural network gave a reasonable accuracy, however, was more difficult to tune than the other methods. The committees implemented, only slightly increased the accuracy obtained from that of the individual neural networks trained. Each method had their advantages and disadvantages in terms of the accuracy obtained, the time required to train, the time required to execute the AI system, the number of parameters to be tuned, and the complexity of the model produced. However, for the prediction of the steam generator data the Adaptive Neuro- Fuzzy System obtained the most accurate predictions.